\documentclass{article}

\usepackage{microtype}
\usepackage{graphicx}

\usepackage{booktabs}

\usepackage{times}
\usepackage{epsfig}
\usepackage{amsmath}
\usepackage{amssymb}
\usepackage{bbm}
\usepackage{textcomp}
\usepackage{caption}
\usepackage{booktabs}
\usepackage{enumitem}
\usepackage{paralist}
\usepackage[hyphens]{url}
\usepackage{graphicx}
\usepackage{subfig}

\usepackage[accepted]{icml2019}
\icmltitlerunning{GEOMetrics: Exploiting Geometric Structure for Graph-Encoded Objects}

\begin{document}

\twocolumn[
\icmltitle{GEOMetrics: Exploiting Geometric Structure for Graph-Encoded Objects}

\icmlsetsymbol{equal}{*}

\begin{icmlauthorlist}
\icmlauthor{Edward J. Smith}{mcgill}
\icmlauthor{Scott Fujimoto}{mcgill,mila}
\icmlauthor{Adriana Romero}{fair,mcgill}
\icmlauthor{David Meger}{mcgill}

\end{icmlauthorlist}

\icmlaffiliation{mcgill}{Department of Computer Science, McGill University, Montreal, Canada}
\icmlaffiliation{mila}{Mila Qu\'ebec AI Institute}
\icmlaffiliation{fair}{Facebook AI Research}%

\icmlcorrespondingauthor{Edward Smith}{edward.smith@mail.mcgill.ca}

\icmlkeywords{Machine Learning, ICML}

\vskip 0.3in
]
\printAffiliationsAndNotice{}

\begin{abstract} 
Mesh models are a promising approach for encoding the structure of 3D objects. Current mesh reconstruction systems predict uniformly distributed vertex locations of a predetermined graph through a series of graph convolutions, leading to compromises with respect to performance or resolution. In this paper, we argue that the graph representation of geometric objects allows for additional structure, which should be leveraged for enhanced reconstruction. Thus, we propose a system which properly benefits from the advantages of the geometric structure of graph-encoded objects by introducing (1) a graph convolutional update preserving vertex information; (2) an adaptive splitting heuristic allowing detail to emerge; and (3) a training objective operating both on the local surfaces defined by vertices as well as the global structure defined by the mesh. Our proposed method is evaluated on the task of 3D object reconstruction from images with the ShapeNet dataset, where we demonstrate state of the art performance, both visually and numerically, while having far smaller space requirements by generating adaptive meshes. 

\end{abstract}

\begin{figure}[t!]
\centering
\captionsetup{justification=centering}

\begin{tabular}{cc}
\subfloat[Voxels \newline ($262,144$ units)]{\includegraphics[width=0.4\linewidth]{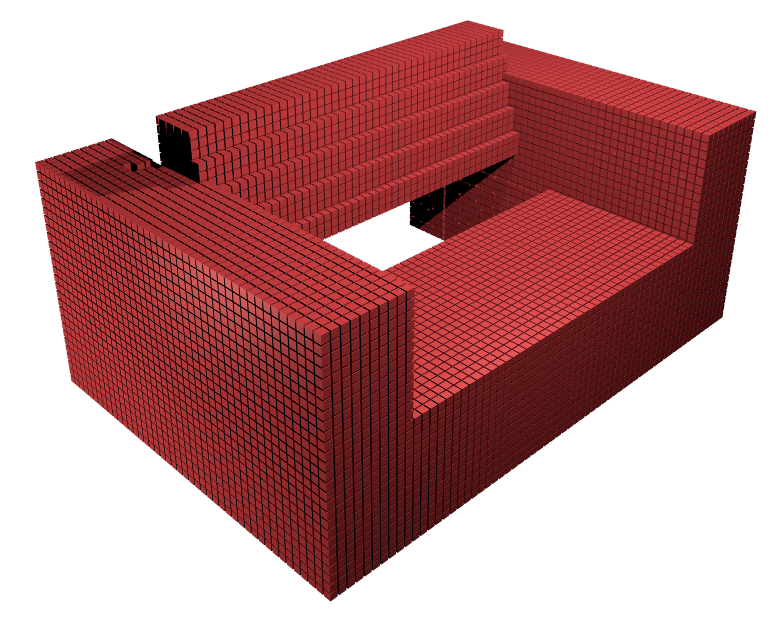} \label{fig:voxels}} &
\subfloat[Point cloud \newline ($30,000$ points)]{\includegraphics[width=0.4\linewidth]{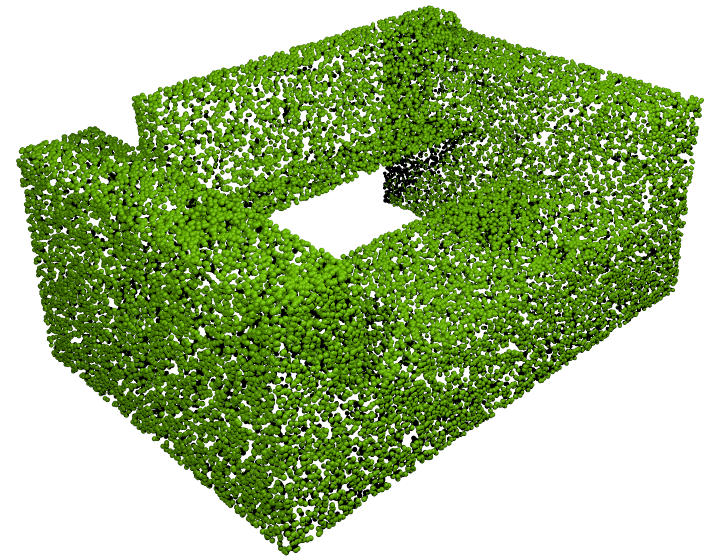} \label{fig:ptcloud}} \\
\subfloat[Uniform mesh \newline ($2416$ vertices)]{\includegraphics[width=0.4\linewidth]{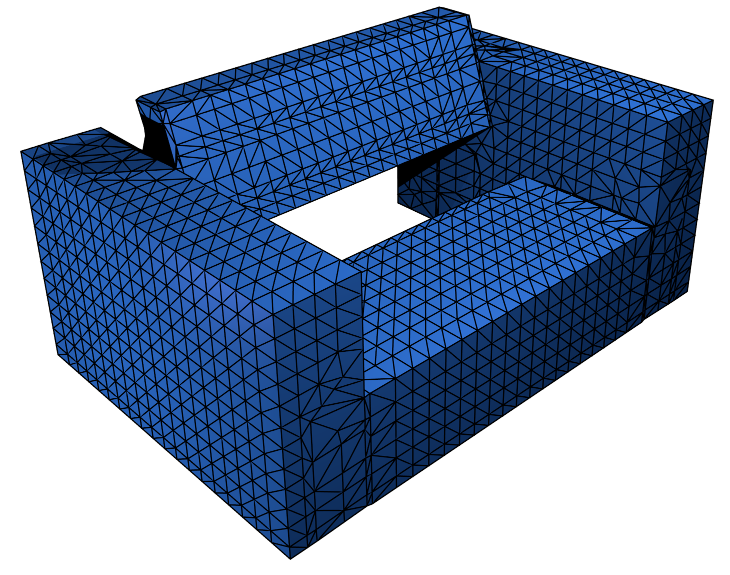}\label{fig:unimesh}} &
\subfloat[Adaptive mesh \newline ($120$ vertices)]{\includegraphics[width=0.4\linewidth]{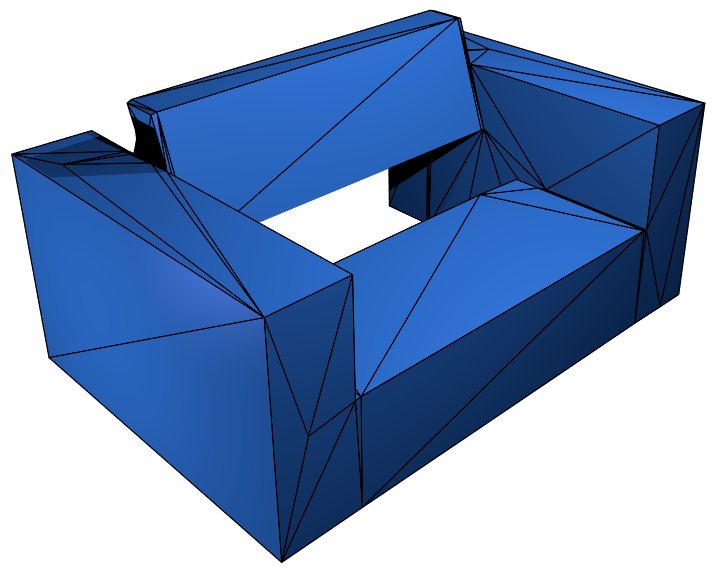}\label{fig:adaptmesh}}
\end{tabular}
\caption{Comparison of 3D shape encoding techniques, including their respective encoding sizes for the level of quality viewed.}
\vspace{-0.3cm}
\label{fig:encoding3D}
\end{figure}

\section{Introduction}
\label{sec:intro}

Surfaces in our physical world exhibit highly non-uniform curvature; compare a plane's wing to a microscopic screw that fastens its engine. Traditionally, deep 3D understanding systems have relied upon representations such as voxels and point clouds, which capture structure with either uniform volumetric or surface detail \cite{choy20163d, 3DGAN, fan2017point}. By representing unimportant, or uninteresting, regions with high detail, these systems scale poorly to higher resolutions and object complexity (see Figure \ref{fig:voxels} and \ref{fig:ptcloud}). While efforts have been made to rectify this issue through intermediary representations \cite{OGN, HSP, mineNIPS, NIPS2018_7494}, these either continue to rely on sparse volumetric units, or maintain uniform detail through alternative representations. 

A triangle mesh is a graph-based shape representation that encodes 3D structure through a set of vertices and corresponding planar faces. Recently, advances in deep learning on graphs have enabled mesh-based 3D shape methods \cite{kato2017neural, Pixel2Mesh, kanazawa2018learning, jack2018learning, henderson2018learning, groueix20183d}. However, these approaches produce mesh predictions which uniformly space vertices and faces over their surface, providing no significant improvement over the previously highlighted representations (see Figure \ref{fig:unimesh}) and hence, not exploiting the mesh representation advantages. In particular, by placing many vertices in regions of fine detail while using large triangles to summarize nearly planar regions, one could define \emph{adaptive meshes} (see Figure \ref{fig:adaptmesh}), enabling flexible scaling by effectively localizing complexity in object surfaces, and allowing for the 3D structure of complicated shapes to be encoded with smaller space requirements and higher precision.

Moreover, these deep learning mesh-based approaches rely on Graph Convolutional Networks (GCNs) \cite{GCN,Defferrard:2016:CNN:3157382.3157527,Cucurull2018ConvolutionalNN,Pixel2Mesh}. Although effective in many node/graph classification and regression tasks, we argue that GCNs may be inadequate for understanding, reconstructing or generating 3D structure as they may induce over-smoothing while aggregating neighboring information at vertex level \cite{Li2018DeeperII}. This aggregation bias could in turn lead to a harder learning problem when vital information held at each vertex cannot be derived from its neighbors, and as a direct consequence must not be lost. 

Last, an important question when reconstructing 3D objects is how to define a loss between a prediction and its target. A common approach is to employ the Chamfer Distance over some parametrization of the two surfaces \cite{Barrow:1977:PCC:1622943.1622971, insafutdinov2018unsupervised,Pixel2Mesh, groueix2018atlasnet,pix3d,fan2017point}. However, this loss penalizes the point positions exclusively, and thus, its direct application to mesh vertex positions leads poor accuracy, as no information of the faces they define is provided and the placement of vertices over a surface is, to a large degree, arbitrary. In addition, this local loss function takes no consideration of the global structure of the predicted object, preventing class-specific attributes to emerge and creating global inconsistencies.

Therefore in this paper, we aim to address the above-mentioned limitations by introducing an \emph{adaptive mesh} reconstruction system, called \textbf{G}eometrically \textbf{E}xploited \textbf{O}bject \textbf{M}etrics (GEOMetrics), which properly capitalizes on the advantages and geometric structure of graph-encoded objects. GEOMetrics reformulates graph convolutional layers to prevent vertex smoothing. Moreover, it incorporates an adaptive face splitting heuristic allowing non-uniform detail to emerge. Finally, it introduces a training objective operating both on the local surfaces defined by vertices, via a differentiable sampling procedure, as well as the global structure defined by the graph, through a perceptual loss reminiscent of that of style transfer applications \cite{gatys2016image,johnson2016perceptual}. To the best of our knowledge, our system is the first deep approach to \emph{describing shape as an adaptive mesh}, through advances in geometrically-aware graph operations. We extensively evaluate our system on the task of 3D object reconstruction from single RGB images and show that the interplay of our introduced components encourages mesh reconstructions, which properly localize detail, while maintaining structural consistency. As a result, we are able to obtain mesh predictions which outperform previous methods and have far smaller space requirements.

The contributions of this paper can be summarized as:

\begin{compactitem}
   
    \item We introduce the Zero-Neighbor GCN (0N-GCN), an extension of \citet{GCN}, which allows the \emph{information at each vertex to be maintained}, and as a result better suits the understanding and reconstruction of 3D meshes. 
    \item We present an adaptive face splitting procedure to \emph{encourage local complexity} to emerge when reconstructing meshes, taking advantage of the mesh flexible scaling (see Figure \ref{fig:adaptmesh}). 
    \item We propose a training objective, which operates locally and globally over the surface to produce mesh reconstructions, which are highly accurate and benefit from the graceful scaling of mesh representations. 
   \item We highlight through extensive evaluation the substantial benefits provided by the previous contributions and show, on the task of 3D object reconstruction from single RGB images, that by properly exploiting the meshs' properties and geometry, our GEOMetrics system is able to notably outperform prior methods visually and quantitatively, while requiring far less vertices/faces. 
\end{compactitem}
Note that the above-mentioned contributions are not specific to the reconstruction system nor the chosen task and thus, can be easily adapted to arbitrary mesh problems. Code for our system is publicly available on a GitHub repository, to ensure reproducible experimental comparison.\footnote{\small https://github.com/EdwardSmith1884/GEOMetrics} 

\section{Related Work}

\textbf{3D Mesh Reconstruction.} \enskip
Mesh models have only recently been used in generation and reconstruction tasks due to the challenging nature of their complex definition \cite{Pixel2Mesh}. Recent mesh approaches rely on graph representations of meshes, and use GCNs \cite{GCN} to effectively process them. Our work most closely relates to Neural 3D Mesh Renderer \cite{kato2017neural} and Pixel2Mesh \cite{Pixel2Mesh}, which use deformations of a generic pre-defined input mesh, generally a sphere, to form 3D structures. Similarly, Atlas-Net \cite{groueix2018atlasnet} uses deformations over a set of primitive square faces to form 3D shapes. Conceptually similar, there exists numerous papers using class-specific input meshes which are deformed with respect to the given input image \cite{pontes2017image2mesh,kanazawa2018learning, jack2018learning,henderson2018learning,groueix20183d, Kar2015CategoryspecificOR}. While effective, these approaches require prior knowledge on the target class or access to a model repository. 

\textbf{Graph Convolutional Networks.} \enskip
The great success of convolutional neural networks in numerous image-based tasks \cite{He_identity_mappings,He2017maskrcnn,Huang2016,jegou2017one,Casanova18} has led to increasing efforts to extend deep networks to domains where graph-structured data is ubiquitous.

Early attempts to extend neural networks to deal with arbitrarily
structured graphs relied on recursive neural networks \cite{Frasconi98,Gori05,ScarselliGNN}. Recently, spectral approaches have emerged as an effective alternative which formulates the convolution as an operation on the spectrum of the graph \cite{Henaff2015DeepCN,bruna2014spectral,Bronstein2017GeometricDL,levie2017cayleynets}. Methods operating directly on the graph domain have also been presented. \citet{Defferrard:2016:CNN:3157382.3157527} proposed to approximate the filters using the Chebyshev polynomials applied on the Laplacian operator. This approximation was further simplified by \citet{GCN}. Finally, several works have been introduced exploring well-established deep learning ideas and improving previously reported results \cite{FingerPrint,GraphSAGE,Monti2017GeometricDL,GAT}.

\textbf{3D Object Representation.} \enskip 
Deep learning approaches for understanding 3D shapes have, for a long time, employed voxels as a default 3D object representation \cite{choy20163d, 3DGAN, mine, tulsiani2017multi, marrnet, wu2018learning}. While straightforward to use, voxels induce a cubic computational cost, scaling poorly to higher resolutions and complex objects. Numerous computationally efficient approaches have arisen, such as octree methods \cite{riegler2017octnet, OGN, HSP}, which represent voxel objects with adaptive degrees of detail. 
Most similar to mesh models are point clouds methods, which represent 3D objects through a set of points in 3D space \cite{fan2017point, qi2017pointnet, insafutdinov2018unsupervised, novotny2017learning}. Point clouds represent only the surface information of 3D objects, making them more efficient and scalable. However, as they do not define surface information beyond each point's local neighborhood they must be uniformly sampled over a surface, and so to encode high levels of detail, the sampling density over the entire surface must increase. 

\section{Background}

In this section, we review GCNs \cite{GCN}, a key component for mesh generation and reconstruction systems, and outline how the Chamfer Distance has been previously employed as a loss for mesh reconstruction.

\subsection{Graph Convolutional Networks}
\label{ssec:GCN}
Let $\mathcal{G}$ be a graph with $N$ vertices defined by an adjacency matrix $\mathbf{A} \in \mathbb{R}^{N \times N}$ and a vertex feature matrix $\mathbf{H} \in \mathbb{R}^{N \times F}$, where $F$ denotes the number of features. 
A GCN layer takes as input both $\mathbf{A}$ and $\mathbf{H}$, and produces a new vertex feature matrix $\mathbf{H'}\in \mathbb{R}^{N \times F'}$ with $F'$ features as follows:
\begin{equation}
H' = \sigma( \mathbf{A} \mathbf{H} \mathbf{W}  + \mathbf{b}),
\label{eq:GCN}
\end{equation}
where $\sigma$ is an arbitrary activation function, and $\mathbf{W} \in \mathbb{R}^{F \times F'}$ and $\mathbf{b} \in \mathbb{R}^F$ are the learnable weight matrix and bias vector, respectively. By stacking multiple such layers, information is exchanged throughout the graph such that, after $k$ layers, the information at a given vertex will, for the first time, reach it's $k^{th}$ depth neighbor. Alternatively, $k^{th}$ depth information can be immediately reached using the $k^{th}$ power of adjacency matrix in Eq.~\ref{eq:GCN} \cite{Defferrard:2016:CNN:3157382.3157527, levie2017cayleynets, Cucurull2018ConvolutionalNN}. Note that, as discussed in Section \ref{sec:intro}, GCN layers equate any given vertex to a summary of its neighbourhood, and their repeated application may over-smooth important local information \cite{Li2018DeeperII}.

\subsection{Chamfer Loss: Vertex-To-Point Loss}

The Chamfer Distance between predicted and ground truth objects has become a standard metric for 3D reconstruction \cite{Pixel2Mesh, insafutdinov2018unsupervised, groueix2018atlasnet,pix3d,fan2017point}. This loss is defined as:
\begin{equation}
\mathcal{L}_{\text{Chamfer}}= \sum_{p \in S} \min_{q \in \hat S} \Vert p - q \Vert^2_2 + \sum_{q \in \hat S} \min_{p \in S} \Vert p - q \Vert^2_2
\label{eq:chamfer}
\end{equation}
and is computed between two sets of points, $\hat S$ and $S$, sampled from the predicted surface and the ground truth surface. As discussed in Section \ref{sec:intro}, this metric performs poorly when directly applied to two sets of mesh vertices as it does not take into account the faces which they define, and because of the difficult learning problem associated with matching highly arbitrary vertex placement on a surface. To avoid these issues, \citet{Pixel2Mesh} define $\hat S$ as a large set of predicted vertex positions, and $S$ as a large, pre-computed set of points sampled from the ground truth surface (see Figure \ref{fig:LossComparison_a}). 
Defining the ground truth set as a dense uniform sampling over the target surface avoids issues with inconsistent vertex positions across similar objects. However, this leads to predictions with a high number of vertices packed tightly over the full surface. In this way, the mesh predictions resemble a point cloud, and thus fail to take advantage of the graceful scaling properties of their representation.

\begin{figure*} 
\centering
\includegraphics[width=1\linewidth, trim={0cm 0cm 0cm 0.0cm}, clip]{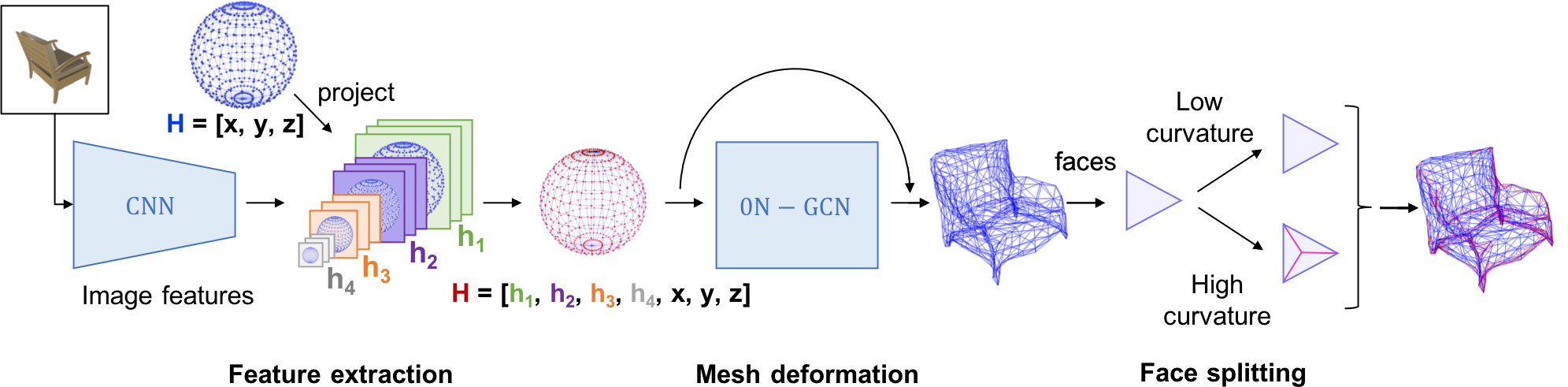}
\vspace{-4mm}
\caption{Mesh reconstruction module, with its three main components highlighted. Feature Extraction describes the process through which image features are extracted for each vertex. Mesh Deformation outlines the deformation of the inputted mesh through 0N-GCN layers. Adaptive Face Splitting illustrates how high curvature faces are split to increase local complexity.} \label{fig:MeshGeneration}
\end{figure*}

\section{GEOMetrics Mesh Reconstruction}

In this section, we describe our pipeline for reconstructing adaptive meshes from single images and outline our proposed 0N-GCN as well as our suggested adaptive face splitting. 

\subsection{Mesh Reconstruction from Images Pipeline} 

Figure \ref{fig:MeshGeneration} depicts our mesh reconstruction module, which takes as input a mesh model, defined by a set of vertex positions and an adjacency matrix, together with an RGB image depicting an object view and outputs a new mesh prediction. The module is composed of three distinct phases: feature extraction, mesh deformation and face splitting, which are cascaded $m=3$ times to obtain incrementally refined mesh predictions. Note that the initial module takes as input a predefined mesh model (e.g. a sphere), whereas each subsequent module is fed the preceding module's prediction. In this manner, the initial mesh is iteratively deformed and updated to match the input image.

Our \emph{feature extraction} is based on the method proposed by \citet{Pixel2Mesh}, where the input image is passed through a deep CNN and the features from 4 intermediary layers are outputted. The feature vector for each vertex of the mesh is then defined by projecting the vertices of the input mesh onto the CNN outputs and extracting their corresponding features. In addition, each vertex feature vector is provided its 3D coordinates $x, y, z$ and also, if available, the final feature vector it possessed in the preceding reconstruction module.
Our \emph{mesh deformation} consists of a graph convolutional model, which takes as input a mesh and deforms it by making a residual prediction for the position of each vertex. The residual prediction is then added to the original position to complete the deformation. The graph convolutional model of this mesh deformation phase is made up of a series of the proposed 0N-GCN layers (see Subsection \ref{ssec:0nGCN} for details). Finally, our \emph{face splitting} phase, described in Subsection \ref{ssec:facesplit}, encourages local complexity to emerge in regions that require additional detail.

\subsection{Zero-Neighbor Graph Convolutional Networks}
\label{ssec:0nGCN}
As described in Subsection \ref{ssec:GCN}, a potential shortcoming of the standard GCN formulation is that a vertex has no capacity to maintain and directly draw conclusions upon its own information, as this information is smoothed with outside influence at each layer. This outside influence, while useful in global graph understanding contexts, may be detrimental when vital information held at each vertex cannot be derived from its neighbors. This situation is exemplified by meshes, where, if optimally defined, every vertex defines some new surface structure (see Figure \ref{fig:adaptmesh}). To rectify this problem, we define a Zero-Neighbor update, in which a fraction of a vertex's feature vector are not updated with the neighbors' information. This is accomplished by, instead of applying higher powers of the adjacency matrix to reach further depths, taking the adjacency matrix to the power $0$ (equivalent to the identity matrix) to exchange with no further depths:
\begin{equation}
\begin{split}
\mathbf{H'} = \mathbf{H}\mathbf{W}, \quad
\mathbf{H''} = \sigma( [\mathbf{A} \mathbf{H'}_{0:i} \Vert \mathbf{A}^0 \mathbf{H'}_{i:}] + \mathbf{b}),
\end{split}
\end{equation}
where $[ \cdot || \cdot]$ denotes concatenation between vectors and $i$ is a feature index. 
This 0N-GCN provides a soft middle ground between full exchange of information and no vertex communication, where the network can choose how heavily a portion of the features of a given vertex will be influenced by the rest of the graph.

\subsection{Adaptive Face Splitting}
\label{ssec:facesplit}

In the final step of each reconstruction module, the mesh's set of vertices is redefined over its surface, by adding vertices in regions of high detail. To do so, we introduce a face splitting method, which adaptively increases the set of vertices and the connections between them by analyzing the local curvature of the surface at each mesh face. 
The curvature at each face is computed by taking the average of the angle between a face's normal and its neighboring faces' normals. For a given face $f$, made up of vertices $v_1, v_2, v_3$, its face normal $N_f$ is calculated as:
\begin{equation}
N_f = \frac{e_1 \times e_2}{\Vert e_1 \times e_2 \Vert},
\end{equation}
where $e_1 = v_1-v_2$ and $e_2 = v_3-v_2$. The curvature $C_f$ at face $f$ is then computed as: 
\begin{equation}
C_f = \frac{180}{|\mathcal{H}_f|\pi} \sum_{i \in \mathcal{H}_f} \arccos{(N_{f} \cdot N_{i})},
\end{equation}
where $\mathcal{H}_f$ is the set of neighboring faces of $f$.
All faces with curvature over a given threshold, $\alpha$, are then selected to be updated. A selected face is updated by adding a new vertex to its center and connecting it to its 3 original vertices, creating 3 new faces. As the new vertex positions are defined by the positions of already existing vertices, the gradients from each vertex are easily defined to flow back through all previous modules. In this way, each reconstruction module is able to identify areas of the current mesh which require increased detail and prescribe them a higher vertex density. This allows the mesh to fully take advantage of the scaling properties of its representation, by concentrating the generation process in areas of high detail.

\begin{figure}[t!]
\subfloat[Vertex-to-point]{\includegraphics[width=0.33\linewidth]{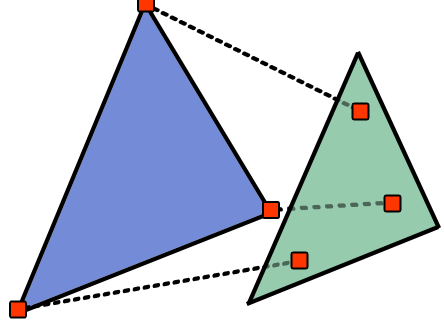} \label{fig:LossComparison_a}}
\subfloat[Point-to-point]{\includegraphics[width=0.33\linewidth]{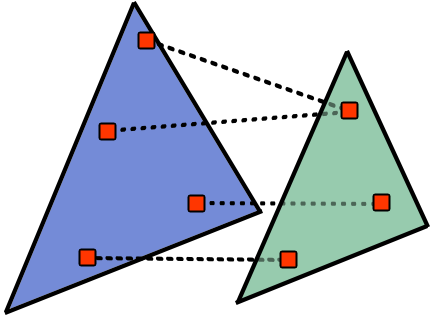} \label{fig:LossComparison_b}}
\subfloat[Point-to-surface]{\includegraphics[width=0.33\linewidth]{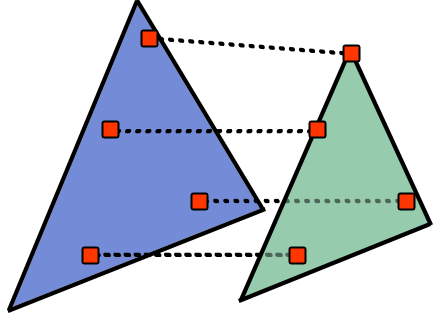} \label{fig:LossComparison_c}}
\caption{A comparison of different surface losses. Vertex-to-point is the technique used by \citet{Pixel2Mesh}. Point-to-point sampling and point-to-surface sampling are the sampling procedures introduced in our approach.}
\label{fig:LossComparison}
\end{figure}

\section{GEOMetrics Losses}

In this section, we describe the key contributions made to the mesh prediction task when considering 3D geometry. In particular, we introduce a training objective, considering the local topology and the global structure to produce mesh predictions that properly benefit from the graph representation.

\subsection{Differentiable Surface Sampling Losses}

We introduce a differentiable sampling procedure which enables us to penalize vertices by the surface they implicitly define, rather then their explicit position. This approach allows predicted meshes to match the target surface without emulating the target vertex positions, which are entirely arbitrary when randomly sampled from the ground truth, while also optimally positioning their vertices and faces. 

To do so, we define a discrete probability distribution based on the relative area of each face and sample $n$ times from this distribution to determine the number of points to sample per face. Then, we sample the previously chosen number of points uniformly over each corresponding surface.More precisely, given a triangular face defined by vertices $v_1, v_2, v_3$, following \citet{osada2002shape}, a point $r$ can be sampled uniformly from the surface of the triangle as:
\begin{equation}
r = (1-\sqrt u)v_1 + \sqrt u(1-w)v_2 + \sqrt{u}wv_3,
\end{equation}
where $u,w \sim U(0,1)$. This formulation allows us to differentiate through the random selection via the reparametrization trick \cite{kingma2013auto, rezende2014stochastic}, as the sampling procedure is defined by a deterministic transformation on the vertex coordinates and the independent stochastic terms.

\begin{figure} [t!]
\includegraphics[width=.9\linewidth]{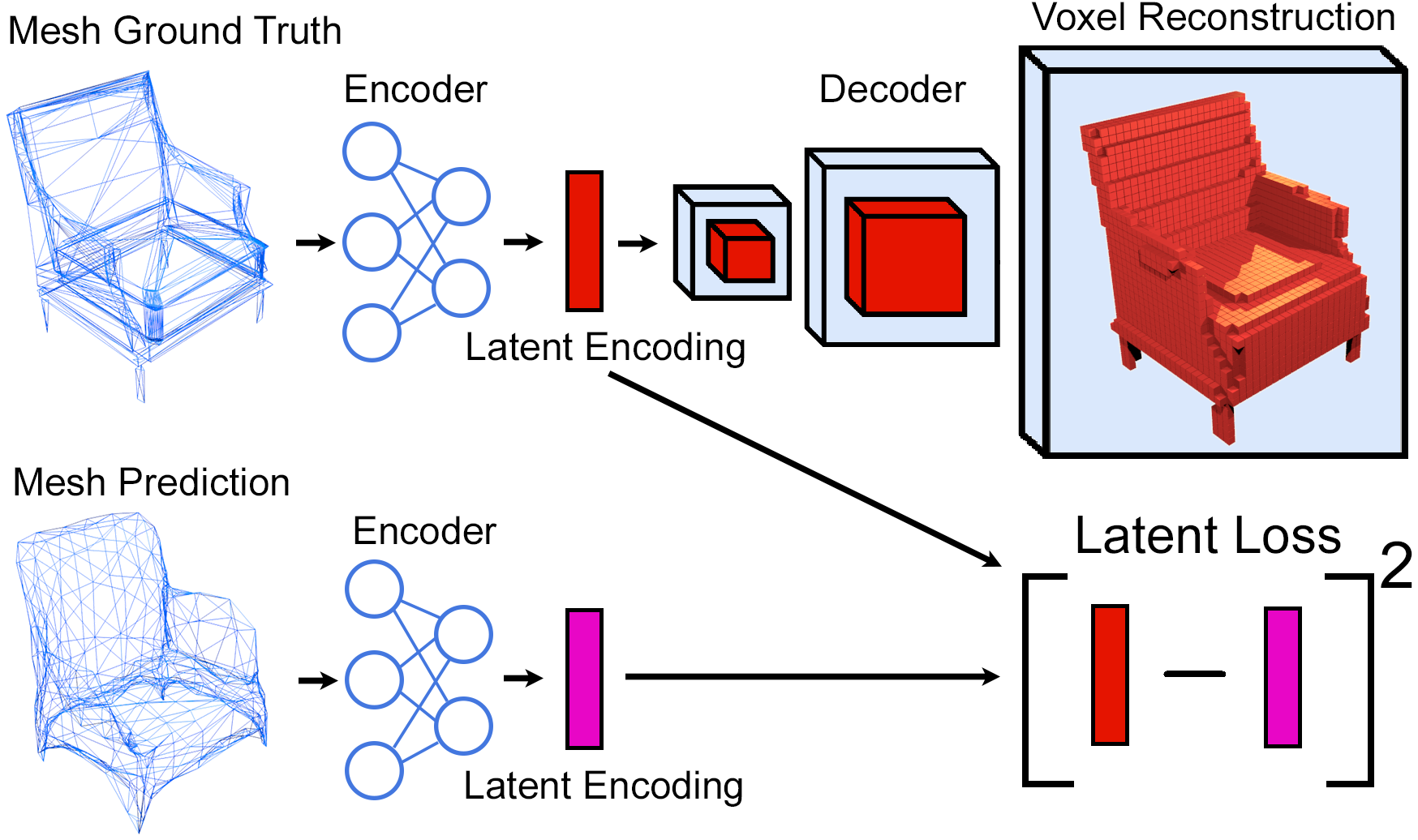}
\centering
\caption{Mesh-to-voxel Mapping. A encoder-decoder architecture is trained to map ground truth meshes to their corresponding voxelizations. The latent representations produced by the encoder from predicted and ground truth meshes are then compared in our global reconstruction loss.} \label{fig:EncoderDecoder}
\end{figure}

We apply this sampling procedure to both the predicted mesh and the ground truth and define an alternative Chamfer loss, \emph{which operates over the previously sampled points, rather than the predicted vertices} (see Figure \ref{fig:LossComparison_b}):
\begin{equation}
\mathcal{L}_{\text{PtP}}= \sum_{p \in S} \min_{q \in \hat S} \Vert p - q \Vert^2_2 + \sum_{q \in \hat S} \min_{p \in S} \Vert p - q \Vert^2_2,
\end{equation} 
where $\hat S$ and $S$ are the sampled points of the predicted mesh and the ground truth, respectively. An algorithmic description of the entire process is provided in the supplementary material. Note this loss differs from the vertex-to-point loss in that it properly penalizes the surface of the predicted mesh instead of the predicted vertex's positions.

Building on these ideas, we define an improved loss term to more accurately compare the surfaces of two meshes: 
\begin{equation}
\mathcal{L}_{\text{PtS}}= \sum_{p \in S} \min_{\hat{f} \in \hat{M}} dist(p, \hat{f}) + \sum_{q \in \hat{S}} \min_{f \in M} dist(q, f), 
\label{eq:ptp}
\end{equation}
where $\hat M$ and $M$ are the predicted and ground truth meshes, $\hat f$ and $f$ the faces, $\hat S$ and $S$ the set of points sampled from the surfaces of $\hat M$ and $M$, and $dist$ is a function computing the distance between a point and a triangular face\footnote{Calculated using an optimized adaptation of the Distance Between Point and Triangle in 3D algorithm \cite{Point2Tri}, details provided in the supplementary material.}. This loss is shown in Figure \ref{fig:LossComparison_c}, and an algorithmic description can be found in the supplementary material. Note that this loss provides an exact measure of the distance between a point and a mesh surface. This is in contrast to the Chamfer loss in Eq.~\ref{eq:chamfer} and the point-to-point loss in Eq. \ref{eq:ptp}, which are faster to compute, but can drastically lose accuracy if too few points are sampled on either surface. A quantitative analysis of the improvement from these losses is provided in the supplementary material through a toy problem.

\begin{figure}[t!]
\centering
\includegraphics[width=.9\linewidth]{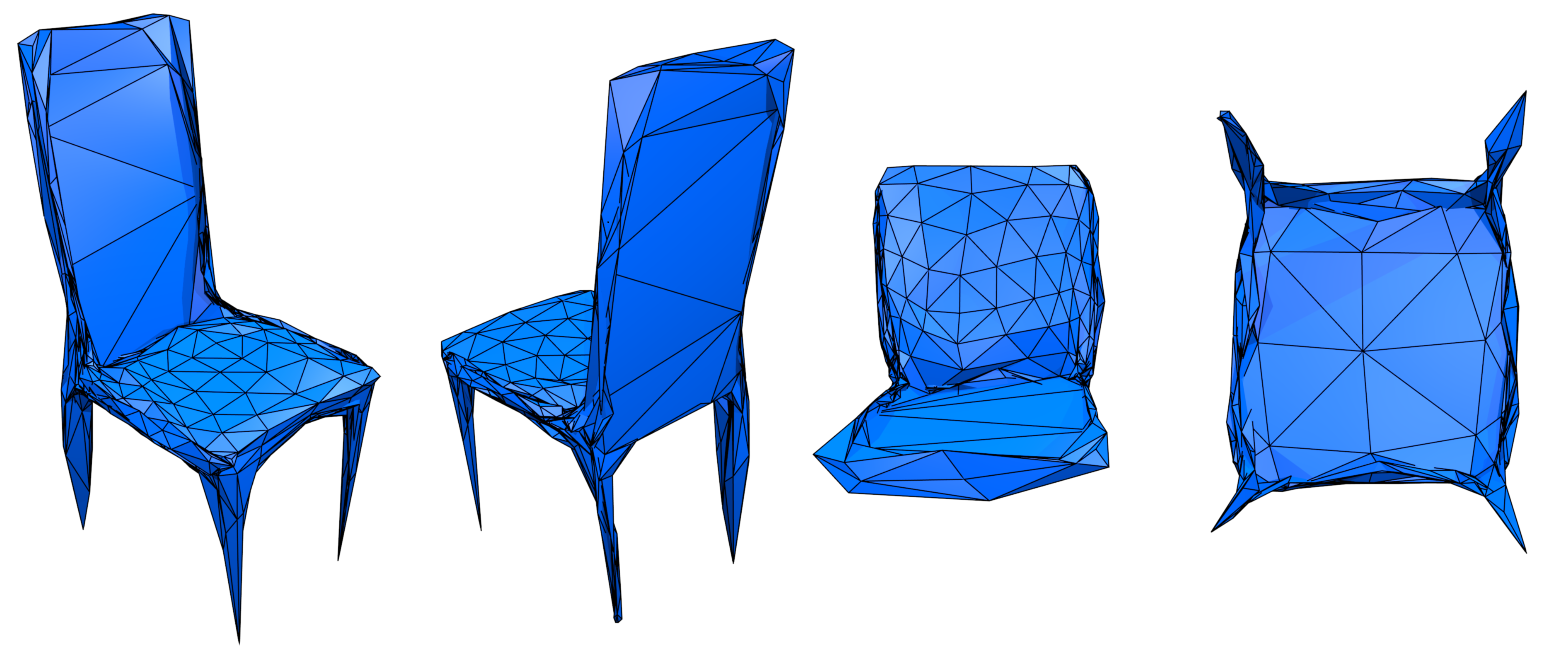}
\vspace{-4mm}
\caption{Renderings from a single reconstructed chair, demonstrating the variety of different local vertex densities which are produced by our approach. } \label{fig:VertexDensity}
\end{figure}

\subsection{Global Encoding of Graphs}
In order to consider the global structure of an object during the reconstruction process, we introduce a global mesh loss. This loss relies on features extracted from a pre-trained mesh-to-voxel model, which is designed as an encoder-decoder network. The mesh-to-voxel encoder takes as input a mesh graph and produces a latent embedding, from which the 3D object is reconstructed, through the decoder, in a voxelized format. In this manner, objects with structural similarity in voxel space, will have similar latent representations, without requiring similar placement of vertices. The proposed global mesh loss is then defined as
\begin{equation}
\begin{split}
 \mathcal{L}_{\text{Latent}} &= || E(M) - E(\hat M)||^2_2 
\end{split}
\end{equation}
where $E$ corresponds to the encoder function of the mesh-to-voxel network. This process is depicted in  Figure \ref{fig:EncoderDecoder}. 

The encoder network of the mesh-to-voxel model is built by stacking 0N-GCN layers, followed by a max pooling operation applied to the set of vertices as in \citet{maxpooling} to produce a single fixed length latent representation. The decoder is a 3D deconvolutional network \cite{choy20163d}, in following with the network defined in \citet{mine}, to perform image to voxel mappings. The complete mesh-to-voxel network is pre-trained by minimizing the mean squared error on the voxelized representations prior to being used in the GEOMetrics system.

\begin{figure}[t!]
\captionsetup[subfloat]{captionskip=0pt}
\centering
\subfloat[Input Image]{\includegraphics[width=0.25\linewidth]{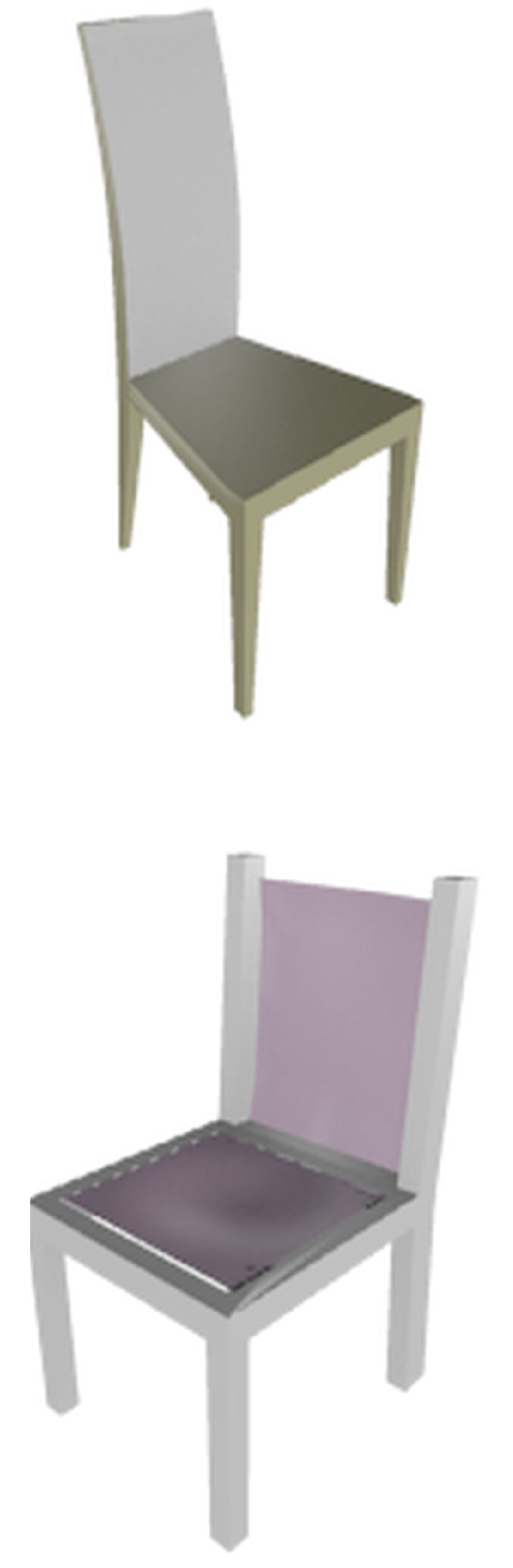} \label{fig:Pixel_input}}\quad
\subfloat[GEOMetrics]{\includegraphics[width=0.25\linewidth]{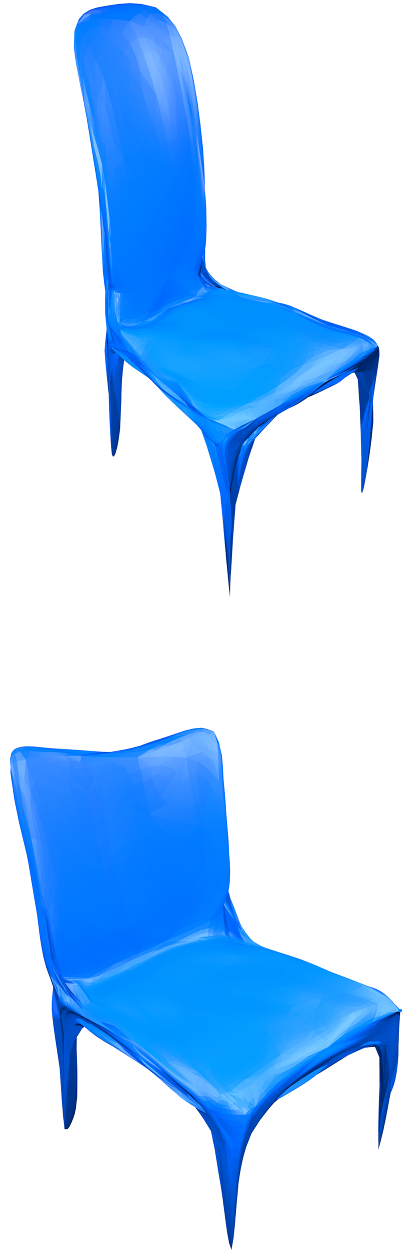} \label{fig:Pixel_GEO}}\quad
\subfloat[Pixel2Mesh]{\includegraphics[width=0.25\linewidth]{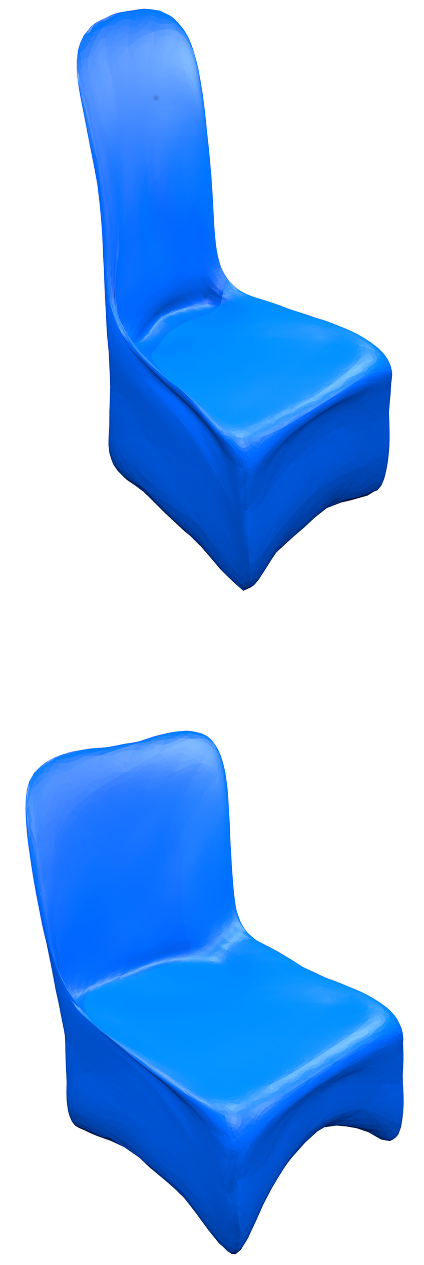} \label{fig:Pixel_Mesh}}
\caption{Visual comparison between GEOMetrics and Pixel2Mesh \cite{Pixel2Mesh} chair reconstructions.}
\label{fig:Pixel_compare}

\end{figure}

\subsection{Optimization Details}

Finally, we present the complete training objective for our mesh reconstruction system. This function combines our differential surface sampling losses, our global structure loss, along with two regularization techniques defined in \citet{Pixel2Mesh}: an edge length minimizing regularizer $\mathcal{L}_{\text{Edge}}$ and a Laplacian-maintaining regularizer $\mathcal{L}_{\Delta \lambda}$, pushing the predicted mesh to be smooth and visually appealing.

The final loss function of our system is defined as: 
\begin{equation}
\mathcal{L} = \gamma_1  \mathcal{L}_{\text{Latent}} + \gamma_2  \mathcal{L}_{\text{PtS}} 
+ \gamma_3  \mathcal{L}_{\text{Edge}} + \gamma_4  \mathcal{L}_{\Delta \lambda}, \label{eq:1}          
\end{equation}
where $\gamma_i$ are hyper-parameters weighting the importance of each term. During the initial stages of training we approximate the $\mathcal{L}_{\text{PtS}}$ term using the defined $\mathcal{L}_{\text{PtP}}$ loss function for faster computation. Note that the loss $\mathcal{L}$ is applied to the output of each mesh reconstruction module. 

\section{Experiments}

\begin{table*}
  \centering
  \caption{Results on ShapeNet 3D object reconstruction reported as per class surface sampling F1 scores and mean F1 score.}
  \vspace{0.2cm}
  \label{table:reconsMesh}
  \scalebox{0.8}{
  \begin{tabular}{l|cc|cc|cc|cc}
    \toprule
    Category 	& 3D-R2N2 & PSG & N3MR & Vertices & Pixel2Mesh & Vertices  & Ours & Vertices\\ 
	& \cite{choy20163d} & \cite{fan2017point} &  \cite{kato2017neural}&  & \cite{Pixel2Mesh} &   &  &\\
    \midrule
    Plane 		& 41.46 & 68.20 & 62.10 & 642 & 71.12 & 2466 &   \bf{89.00} & 645.03\\
    Bench 		& 34.09 & 49.29 & 35.84 & 642 & 57.57 & 2466 &   \bf{72.11} & 514.54\\
    Cabinet 	& 49.88 & 39.93 & 21.04 & 642 & \bf{60.39} & 2466 & 59.52 & 556.68\\
  	Car 		& 37.80 & 50.70 & 36.66 & 642 & 67.86 & 2466 &  \bf{74.64} & 509.33\\
    Chair 		& 40.22 & 41.60 & 30.25 & 642 & 54.38 & 2466 &  \bf{56.61} & 619.13\\
  	Monitor 	& 34.38 & 40.53 & 28.77 & 642 & 51.39 & 2466 &  \bf{59.50} & 449.65\\
    Lamp 		& 32.35 & 41.40 & 27.97 & 642 & 48.15 & 2466 &  \bf{58.65} & 743.28\\
    Speaker 	& 45.30 & 32.61 & 19.46 & 642 & 48.84 & 2466 &  \bf{49.53} & 550.06\\
  	Firearm 	& 28.34 & 69.96 & 52.22 & 642 & 73.20 & 2466 &  \bf{88.36} & 638.35\\
    Couch 		& 40.01 & 36.59 & 25.04 & 642 & 51.90 & 2466 &  \bf{59.54} & 561.79\\
  	Table 		& 43.79 & 53.44 & 28.40 & 642 & 66.30 & 2466 &  \bf{66.33} & 732.82 \\
    Cellphone 	& 42.31 & 55.95 & 27.96 & 642 & 70.24 & 2466 &  \bf{73.65} & 416.05\\
  	Watercraft 	& 37.10 & 51.28 & 43.71 & 642 & 55.12 & 2466 &  \bf{68.32} & 526.04\\
    \midrule
    Mean        & 39.01 & 48.58 & 33.80 & 642 & 59.72 & 2466 &  \bf{67.37} & 574.06\\
    \bottomrule
  \end{tabular}}
\end{table*}

In this section, we demonstrate our algorithm's ability to reconstruct the surface information of 3D objects from single RGB images by taking advantage of the benefits of the mesh representation. We evaluate on this task across 13 classes of the ShapeNet \cite{ShapeNet} dataset. In addition, we present an ablation study to demonstrate how our algorithm's individual components contribute to its overall performance. 

\subsection{Dataset}

The dataset consists of mesh models, voxel models, and RGB images computed from 13 large classes of CAD models found in the ShapeNet dataset \cite{ShapeNet}. Mesh models were computed from the CADs by removing all texture information and downscaling their size so that each model possesses less then 2000 vertices, where possible. From these mesh models, voxelized counter parts were produced at $32^3$ resolution. From each CAD model, 24 RGB images were produced, from random viewpoints, with the camera projection matrix recorded for use in the feature selection method. The data in each class was then split into a training, validation and test set with a ratio of 70:10:20, respectively. This matches the dataset used for empirical evaluation by \citet{Pixel2Mesh} and \citet{mineNIPS}.

\subsection{Implementation Details} 
\textbf{Mesh-to-Voxel Mapping} \enskip For each class, we train a mesh-to-voxel mapping from the mesh and voxel ground truths, for use in our latent loss. These mappings are trained with Adam optimizer \cite{kingma2014adam} ($\beta_1$ = 0.9, $\beta_2$ = 0.999), a learning rate of $10^{-4}$, and a mini-batch size of 16. We train for $10^5$ iterations and practice early stopping, with the best model selected from evaluating on the validation set every 100 iterations. 

\textbf{GEOMetrics} \enskip We train the full system on each class in our dataset with Adam optimizer \cite{kingma2014adam}, at learning rate of $10^{-4}$ for $300$k iterations, and then again for $150$k iterations at a learning rate of $10^{-5}$, with mini-batch size of 5. We practice early stopping by evaluating on the validation set every 50 iterations. The hyper-parameter settings used, as described in Eq.~\eqref{eq:1}, are $\gamma_1 = .001$, $\gamma_2 = 1$,  $\gamma_3 = 0.3$, and $\gamma_4 = 1$. As mentioned above,  
$\mathcal{L}_{PtP}$ is employed as a faster approximation to the $\mathcal{L}_{PtS}$ loss, specifically for the first 300k iteration. This is because $\mathcal{L}_{PtS}$ is slow to compute and we found it sufficient to only use it to finetune pre-trained models. All hyper-parameters were initially tuned on the validation set of the chair class.  The generic pre-defined mesh fed to the first reconstruction module is an ellipsoid. A face is split at the end of each module only if the curvature at that face is greater than 70\textdegree. Architecture details for all networks is provided in the supplementary. 

\begin{figure*}[ht!]
\centering
\includegraphics[width=\textwidth]{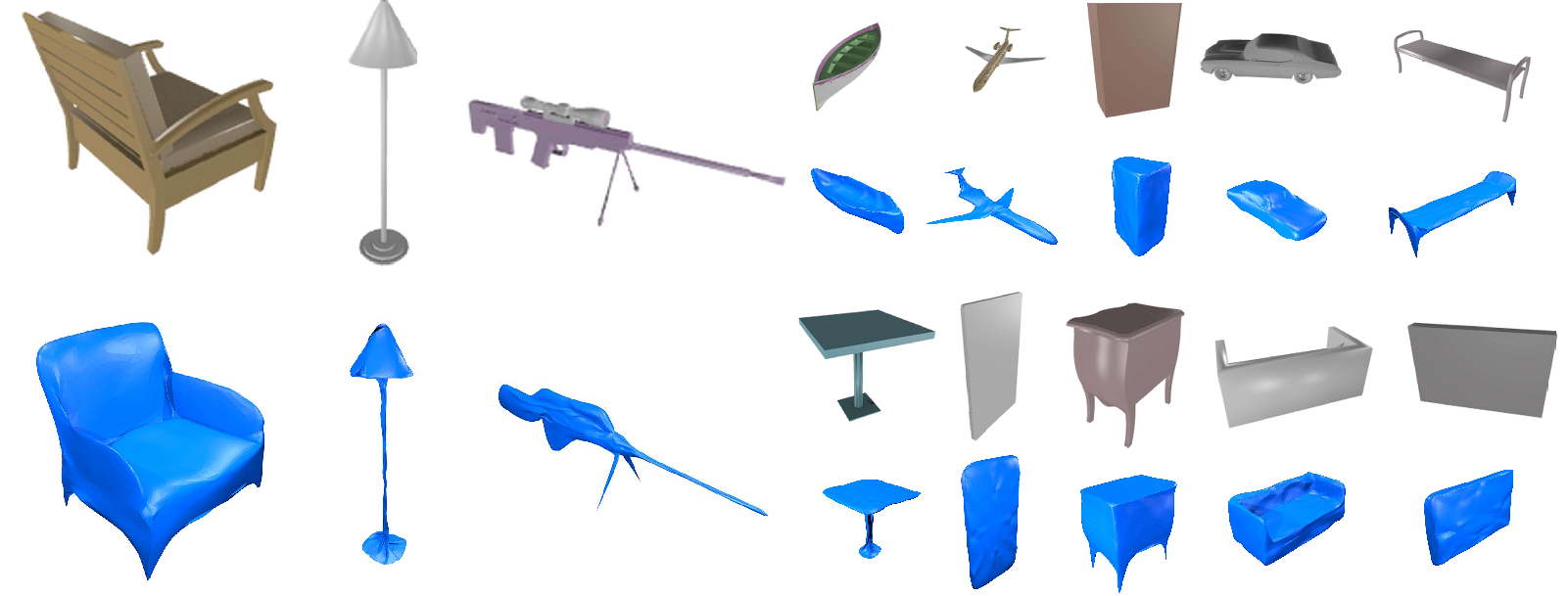}
\vspace{-0.4cm}
\caption{Qualitative results: renderings of meshes reconstructed from each ShapeNet object classes.} \label{fig:ReconstructionResults}
\vspace{-0.2cm}
\end{figure*}

\subsection{Single Image Reconstruction}

We evaluate our method's performance quantitatively by comparing its ability to reconstruct mesh surfaces from single RGB image to an array of high performing 3D object reconstruction algorithms. To do this, we sample points from both the surface of the predicted object and the ground-truth object and compute the F1 score. In following with \cite{Pixel2Mesh}, precision and recall are calculated using the percentage of sampled points which exists within a threshold of $0.0001$ any sampled point in the compared surface. State of the art results of mesh approaches, N3MR \cite{kato2017neural} and Pixel2Mesh \cite{Pixel2Mesh}, a point cloud method, PSG \cite{fan2017point} and a voxel baseline, 3D-R2N2 \cite{choy20163d}, are reported from \citet{Pixel2Mesh}. We also compare mesh-based approaches in terms of space requirements (number of vertices). The results of this comparison are summarized in Table \ref{table:reconsMesh}. As shown, our GEOMetrics system boasts far higher performance than previous approaches, with an average increase in F1 score of $7.65$ points across all classes, and improved score in all classes but one, where we experience a negligible drop of $0.87$ points. In addition, in all cases, our system requires notably less vertices than the previous mesh-based state of the art Pixel2Mesh, e.g. cellphone objects require $6\times$ less vertices, whereas lamp objects require $3\times$ less vertices. With an average of $574.06$ vertices used across all classes, the vertex requirements drop as much as $4.3\times$ on average, highlighting the potential of the adaptive face splitting. Moreover, when compared to point cloud and voxel baselines, we also exhibits state of the art results.

Qualitative reconstruction results for each of the 13 classes are displayed in Figure \ref{fig:ReconstructionResults}\footnote{See supplementary material for additional visualizations.}. We boast highly accurate reconstructions of the input object, effectively capturing both global structure and local detail. In addition, we render an un-smoothed chair reconstruction in Figure \ref{fig:VertexDensity} with its edges heavily outlined, demonstrating the obtained diverse vertex density across a single object and highlighting the way our system represents simple surfaces with a small number of faces, and shifts to higher density where required. Lastly, Figure \ref{fig:Pixel_compare} depicts a visual comparison between GEOMetrics and Pixel2Mesh reconstructions, where we can observe how GEOMetrics is able to provide reconstructions with higher detail (e.g. the sharpness of the chair legs).

\subsection{Single Image Reconstruction Ablation Study}

In this subsection, we study the influence of our system's components and demonstrate their individual importance by comparing our full method's results on the chair class to ablated versions of the system. We assess the impact of our 0N-GCN layers by replacing them with standard GCNs \cite{GCN} in both the mesh reconstruction as well as the mesh-to-voxel models. We validate the effectiveness of the proposed adaptive face splitting by substituting it by a procedure in which all faces are split at the end of each reconstruction module, keeping uniform detail and maintaining approximately the same the number of vertices as the full approach. We then check the importance of each one of the newly introduced losses by removing them at training time. Note that, when the face sampling losses $\mathcal{L}_{\text{PtP}}$ and $\mathcal{L}_{\text{PtS}}$ are removed, we replace them by the vertex-to-point loss ($\mathcal{L}_{\text{VtP}}$) proposed by \citet{Pixel2Mesh}. Finally, we compare our method to the Pixel2Mesh, when roughly equivalent in terms of number of vertices.

The results of this ablation study are reported in Table \ref{table:ablation}. As shown in the table, the biggest effect comes from the introduction of the adaptive face splitting, with a drop of $6.23$ points when replacing it with a uniform splitting heuristic. Moreover, assisting the model to give more importance to vertex features through the 0N-GCN also appears to be relevant. The losses proposed to train the whole system also play an important role, as ignoring them leads to a decrease in performance of $3.69$ points and $1.02$ points, respectively. Moreover, training Pixel2Mesh baseline to use as few vertices as GEOMetrics leads to notably worse performance. These results empirically justify the contributions of our GEOMetrics system.

\begin{table} [H]
  \caption{GEOMetrics ablation compared to full method (ours) and Pixel2Mesh. Results reported as mean F1 score on the chair class.
} 
\vspace{0.2cm}
  \centering
  \scalebox{0.83}{
  \begin{tabular}{c|cccc|c}
    \toprule
    Ours & GCN &  Unif. Split. & No $\mathcal{L}_{latent}$ &  $\mathcal{L}_{\text{VtP}}$ & Pixel2Mesh\\ 
    \midrule 
    \bf{56.61} & 54.57 & 50.33 & 55.59 & 52.92 & 38.13\\
    \bottomrule
  \end{tabular}}
  \label{table:ablation}
\end{table}
\vspace{-3mm}

\section{Conclusion}

In this paper, we presented GEOMetrics, a novel approach for adaptive mesh reconstruction, which focuses on exploiting the geometry of the mesh representation. The GEOMetrics system reformulates GCNs to explicitly preserve local vertex information and incorporates an %
adaptive face splitting procedure to enhance local complexity when necessary. Furthermore, the system is trained by introducing a training objective which operates both locally and globally at mesh level, and capitalizes on the geometric structure of graph-encoded objects. We demonstrated the potential of the approach through extensive evaluation on the challenging task of 3D object reconstruction from single images of the ShapeNet dataset. Finally, we reported visually appealing state of the art results, outperforming existing mesh-based methods by a large margin, while requiring (on average) as many as $4.3\times$ less vertices. %
Future research directions include addressing the restrictive constant topology prescribed by the initial mesh object through reconstruction and generation methods, which adapt the topology to match the target mesh.

\bibliographystyle{icml2019}

\clearpage

\twocolumn[\icmltitle{GEOMetrics: Supplemental Material}]
\appendix

\section{Point to Surface Loss}

In this section, we describe the the Distance Between Point and Triangle in 3D algorithm \cite{Point2Tri}. 
For a given point $P$ and triangle $T$, the algorithm computes the minimum distance between the point and any point contained within the triangle. Assuming the triangle is defined by corner point $B$ and directions $E_0$ and $E_1$, then any point $T(s,t)$ contained in the triangle can be defined by a pair of scalars $(s,t)$ such that $T(s,t) = B +sE_0 + tE_1$, where $(s, t) \in D = \left\{ (s,t) :s \geq 0 , t \geq 0, s+t \leq 1 \right\}$. We can now define the squared distance $Q$ between the point $P$ and any point in the triangle $T(s,t)$ by the following quadratic function: 
\begin{equation}
Q(s,t) =  as^2 + 2bst + ct^2 + 2ds + 2et + f,
\end{equation}
where for clarity we denote $a = E_0 \cdot E_0$ , $b = E_0 \cdot E_1$, $c = E_1 \cdot E_1$, $d = E_0 \cdot (B - P)$, $e = E_1 \cdot (B - P)$, and $f = (B - P) \cdot (B - P)$. Selecting $(s,t)$ which minimizes $Q(s,t)$ provides the minimum distance between the point $P$ and triangle $T$. As $Q$ is continuously differentiable, $(s,t)$ can be found at an interior point where $\nabla Q = 0$ or at the boundary of the set $D$. 

In the first case, note $\nabla Q(s',t') = 0$ if and only if $s'$ and $t'$ satisfy the following:
\begin{equation}
s' = \frac{be - cd}{ac-b^2}, \qquad t' = \frac{bd - ae}{ac-b^2}
\end{equation}
Then if $(s',t') \in D$, we have the minimum distance.Otherwise, the distance minimizing $(s,t)$ must lie on the boundary of D, where either $\left\{s=0, t \in [0,1]\right\}$, $\left\{s\in[0,1], t = 0 \right\}$, or $\left\{s\in[0,1], t = 1-s\right\}$. In each case $Q(s,t)$, can be reduced to quadratic of one unknown variable, which can be minimized by setting the gradient to $0$.

\section{Mesh-to-Voxel Mapping Ablation}

In this section, we perform an ablation study over the use of 0N-GCN as building block for our Mesh-to-Voxel Mapping network to highlight its impact with respect to the standard GCN layers. To that end, we compare our model on 3 different object classes to an analogous network composed of standard GCN layers with the same number of parameters. Additionally, we assess the influence of pooling across a set of vertices by comparing it to other forms of aggregation such as the one introduced by the Neural Graph Fingerprint (NGF) model \cite{FingerPrint}. The results of this ablation study can be found in Table \ref{table:meshtovoxel} in terms of mean squared error (MSE). As shown in the table, results demonstrate the benefits of the 0N-GCN layers, as well as the max-pooling vertex set aggregation, for this mesh understanding task.

\begin{table}[h]
  
  \caption{Mesh-to-Voxel Mapping Reconstruction MSE scores.} \label{table:meshtovoxel}
  \vspace{0.2cm}
  \centering
  \scalebox{0.9}{
  \begin{tabular}{l|cccc}
    \toprule
    Category 	&  Ours & GCN & NGF \\ 
    \midrule
    Plane 		& \bf{0.0089} & 0.0104 & 0.0108 \\
   	table 		& \bf{0.0310} & 0.0393 & 0.0360 \\
    Chair 		& \bf{0.0412} & 0.0526 & 0.0486 \\
    \midrule
    Mean        & \bf{0.0270} & 0.0341 & 0.0318  \\
    \bottomrule
  \end{tabular}}
  
\end{table}

\section{Differentiable Surface Loss Algorithms}

This section provides the algorithmic details of both the point-to-point loss (Algorithm \ref{alg:ptp}) as well as the point-to-surface loss (Algorithm \ref{alg:pts}).

\begin{algorithm}[tbh]
   \caption{Point-to-Point Loss}
   \label{alg:ptp}
\begin{algorithmic}[1]
    \STATE {\bfseries Input:} Two mesh surfaces $M$ and $\hat{M}$, and number of points $n$
    \FOR{face $f$ in mesh $M$}
    \STATE $A_f = Area(f)$
    \STATE $A_T \mathrel{{+}{=}} Area(f)$
    \ENDFOR
    \STATE Define $F \;$ s.t. $P(F = f) =  \frac{A_f*100}{A_T}$
    \STATE Define $U = Uniform(0,1)$
    \STATE Define S = [] 
    \FOR{$i=0$ {\bfseries to} $n$}
    \STATE $f \sim F$
    \STATE $v_1, \; v_2, \; v_3  = vertices(f)$
    \STATE $u \sim U, \; w \sim U $
    \STATE $r = (1-\sqrt u)v_1 + \sqrt u(1-w)v_2 + \sqrt{u}wv_3$
    \STATE $S.append(r)$
    \ENDFOR
    \STATE Apply lines 1 to 15 to mesh $\hat{M}$ to produce $\hat{S}$
    \STATE $\displaystyle \mathcal{L}_{\text{PtP}}= \sum_{p \in S} \min_{q \in \hat S} \Vert p - q \Vert^2_2 + \sum_{q \in \hat S} \min_{p \in S} \Vert p - q \Vert^2_2$
\end{algorithmic}
\end{algorithm}

\begin{algorithm}[tbh]
   \caption{Point-to-Surface Loss}
   \label{alg:pts}
\begin{algorithmic}[1]
    \STATE {\bfseries Input:} Two mesh surfaces $M$ and $\hat{M}$, and number of points $n$
    \FOR{face $f$ in mesh $M$}
    \STATE $A_f = Area(f)$
    \STATE $A_T \mathrel{{+}{=}} Area(f)$
    \ENDFOR
    \STATE Define $F \;$ s.t. $P(F = f) =  \frac{A_f*100}{A_T}$
    \STATE Define $U = Uniform(0,1)$
    \STATE Define S = [] 
    \FOR{$i=0$ {\bfseries to} $n$}
    \STATE $f \sim F$
    \STATE $v_1, \; v_2, \; v_3  = vertices(f)$
    \STATE $u \sim U, \; w \sim U $
    \STATE $r = (1-\sqrt u)v_1 + \sqrt u(1-w)v_2 + \sqrt{u}wv_3$
    \STATE $S.append(r)$
    \ENDFOR
    \STATE Apply lines 1 to 15 to mesh $\hat{M}$ to produce $\hat{S}$
    \STATE $\displaystyle L_{\text{PtS}} = \sum_{p \in S} \min_{\hat{f} \in \hat{M}} dist(p, \hat{f}) + \sum_{q \in \hat{S}} \min_{f \in M} dist(q, f)$
\end{algorithmic}
\end{algorithm}

\section{Loss Analysis}
\begin{figure}[h!]
\centering
\includegraphics[width=.9\linewidth,trim={0 0 0 4mm},clip]{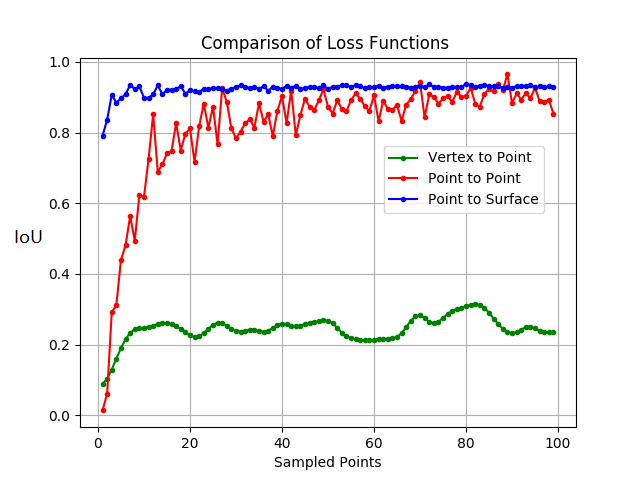}
\vspace{-2mm}
\caption{Comparison of vertex-to-point, point-to-point and surface-to-point losses, in terms of IoU, on a toy problem: optimizing the placement of vertices of a square to match that of a target triangle. Results are compared by increasing the number of sampled points on the surfaces they optimize. Vertex-to-point is the loss employed by \citet{Pixel2Mesh}. Point-to-point and point-to-surface are the losses introduced in our paper. } \label{fig:Loss_graph}
\end{figure}

\begin{figure}[h!]
\centering
\includegraphics[width=0.9\linewidth]{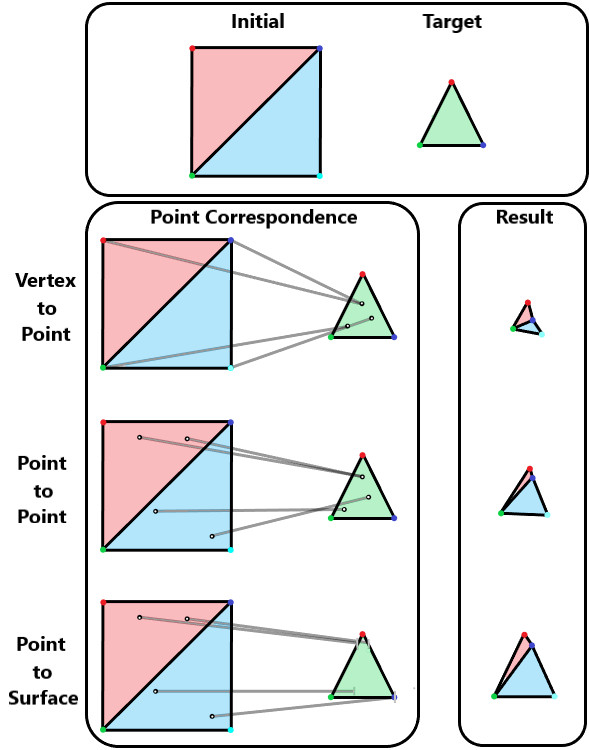}
\vspace{-2mm}
\caption{Qualitative comparison of vertex-to-point \cite{Pixel2Mesh}, point-to-point and surface-to-point losses on the square-to-triangle problem when using $50$ sampled points. We highlight the correspondence between the points in the initial surface and the target surface, which are chosen to be compared, and show the result of optimizing the placement of the vertices when using each of the losses.} \label{fig:SecondLossComparison}
\end{figure}

In this section, we present further analysis of GEOMetrics losses to emphasize the benefits of the introduced point-to-point and surface-to-point losses over the vertex-to-point loss. To that end, we design a toy problem, which consists in optimizing the placement of the vertices of an initial square surface to match the surface area of a target triangle in 2D. Figure \ref{fig:SecondLossComparison} (top) depicts the above-mentioned initial and target surfaces. We optimize the placement of the vertices of the initial square by performing gradient descent on each of the losses independently, and calculate the intersection over union (IoU) of the predicted object and the target triangle. Moreover, in order to assess the impact of the number of points sampled, we repeat this experiment $100$ times, increasing the number of sampled points from $1$ to $100$. Figure \ref{fig:Loss_graph} shows the results of this experiment. Firstly, we observe that the vertex-to-point loss fails to match the target surface entirely, no matter the number of sampled points. Secondly, we observe that the point-to-point loss performance is notably affected by the number of sampled points. While it exhibits poor performance for lower number of sampled points (e.g. below $20$), it rapidly improves as the number of sampled points increases, and ultimately, converges to an average performance, which is only slightly lower than that of the point-to-surface loss. Finally, the point-to-surface loss begins with a far higher IoU and remains the stronger option for nearly all numbers of sampled points.

Figure \ref{fig:SecondLossComparison} illustrates qualitative results for the three compared losses when optimizing with 50 points sampled. As can be seen, the point-to-surface deformation of the square better matches the target triangle shape, followed by point-to-point, which somewhat emulates the triangle, and vertex-to-point, which exhibits the poorest results.

\vspace{-2mm}

\vspace{-2mm}

\clearpage

\section{Network Architectures}

In this section, we provide details on the architectures of the networks used in the paper. Table \ref{table:VGG} describes the feature extractor network of the mesh reconstruction module. Similarly, Table \ref{table:MeshDeform} specifies the mesh deformation network of the reconstruction module. Finally, Table \ref{table:MeshEncoder} and \ref{table:MeshDecoder} detail the mesh-to-voxel encoder and decoder architectures, respectively. 

\begin{table*}[ht]
\centering
\begin{tabular}{|l|c|c|c|c|c|c|c|c|}
\hline
Layers          & 1-2 & 3-8 & 9   & 10-11 & 12  & 13-14 & 15  & 16-18 \\ \hline
Output Resolution    & $224\times224$ & $224\times224$ & $112\times112$   & $112\times112$ & $56\times56$  & $56\times56$ & $28\times28$  & $28\times28$ \\ \hline
\# Channels     & 16  & 32  & 128 & 128   & 256 & 256   & 512 & 512   \\ \hline
Kernel Size     & $3\times3$ & $3\times3$ & $3\times3$ & $3\times3$ & $3\times3$ & $3\times3$ & $3\times3$ & $3\times3$  \\ \hline
Stride     & 1   & 1   & 2   & 1     & 2   & 1     & 2   & 1     \\ \hline
Extracted Layer & -   & 8   & -   & 11    & -   & 14    & -   & 18    \\ \hline
\end{tabular}
\caption{\textbf{Feature extraction network:} Details of the convolutional neural network architecture used to extract image features. Each layer performs a 2D convolutional, followed by batch normalization \cite{ioffe2015batch} and a ReLU activation function \cite{nair2010rectified}. The last row indicates which layer's features are extracted for use in the mesh reconstruction module.}
\label{table:VGG}
\end{table*}

\begin{table*}[ht]
\centering
\begin{tabular}{|l|c|c|c|}
\hline
Layers                       & 1   & 2-14 & 15  \\ \hline
Input Feature Vector Length  & 3   & 192  & 192 \\ \hline
Output Feature Vector Length & 192 & 192  & 3   \\ \hline
\end{tabular}
\caption{\textbf{Mesh deformation network:} Details of the graph convolutional network architecture used to compute the mesh deformation in each reconstruction module. Each layer is composed of a 0N-GCN, followed by an ELU activation function \cite{clevert2015fast}.}
\label{table:MeshDeform}
\end{table*}

\begin{table*}[ht]
\centering
\begin{tabular}{|l|c|c|c|c|c|c|c|c|c|c|c|}
\hline
Layers      & 1      & 2-4    & 5      & 6-7    & 8      & 9      & 10     & 11     & 12     & 13-16  & 17              \\ \hline
Input Feature Dimension  & 3      & 60     & 60     & 120    & 120    & 150    & 200    & 210    & 250    & 300    & 300             \\ \hline
Output Feature Dimension & 60     & 60     & 120    & 120    & 150    & 200    & 210    & 250    & 300    & 300    & 50              \\ \hline

\end{tabular}
\caption{\textbf{Mesh-to-voxel encoder:} Details of the graph convolutional network architecture used to encode mesh graphs into latent vectors. Each layer is composed of a 0N-GCN followed by an ELU activation function \cite{clevert2015fast}, except for the final layer which is a max pooling aggregation over the set of vertices.}
\label{table:MeshEncoder}
\end{table*}

\begin{table*}[ht]
\centering
\begin{tabular}{|l|c|c|c|c|c|}
\hline
Layers             & 1      & 2      & 3        & 4        & 5        \\ \hline
Output Resolution  & $4\times4\times4$  & $8\times8\times8$  & $16\times16\times16$ & $32\times32\times32$ & $32\times32\times32$ \\ \hline
\# Channels & 64     & 64     & 32       & 8        & 1        \\ \hline
Stride      & 2      & 2      & 2        & 2        & 1        \\ \hline
Type               & DeConv & DeConv & DeConv   & DeConv   & Conv     \\ \hline
\end{tabular}
\caption{\textbf{Mesh-to-voxel decoder:} Details of the 3D convolutional neural network archictecture used to decode latent vectors into voxel grids. Each layer performs a 3D deconvolution \cite{shelhamer2017fully} with batch normalization \cite{ioffe2015batch} and an ELU activation function \cite{clevert2015fast}, except for the final layer which is a standard 3D convolutional layer.}
\label{table:MeshDecoder}
\end{table*}

\section{Single Image Reconstruction Visualizations} 
\begin{figure*}[h!]
\centering
\begin{tabular}{ccc}
\subfloat[bench]{\includegraphics[width = 1.5in]{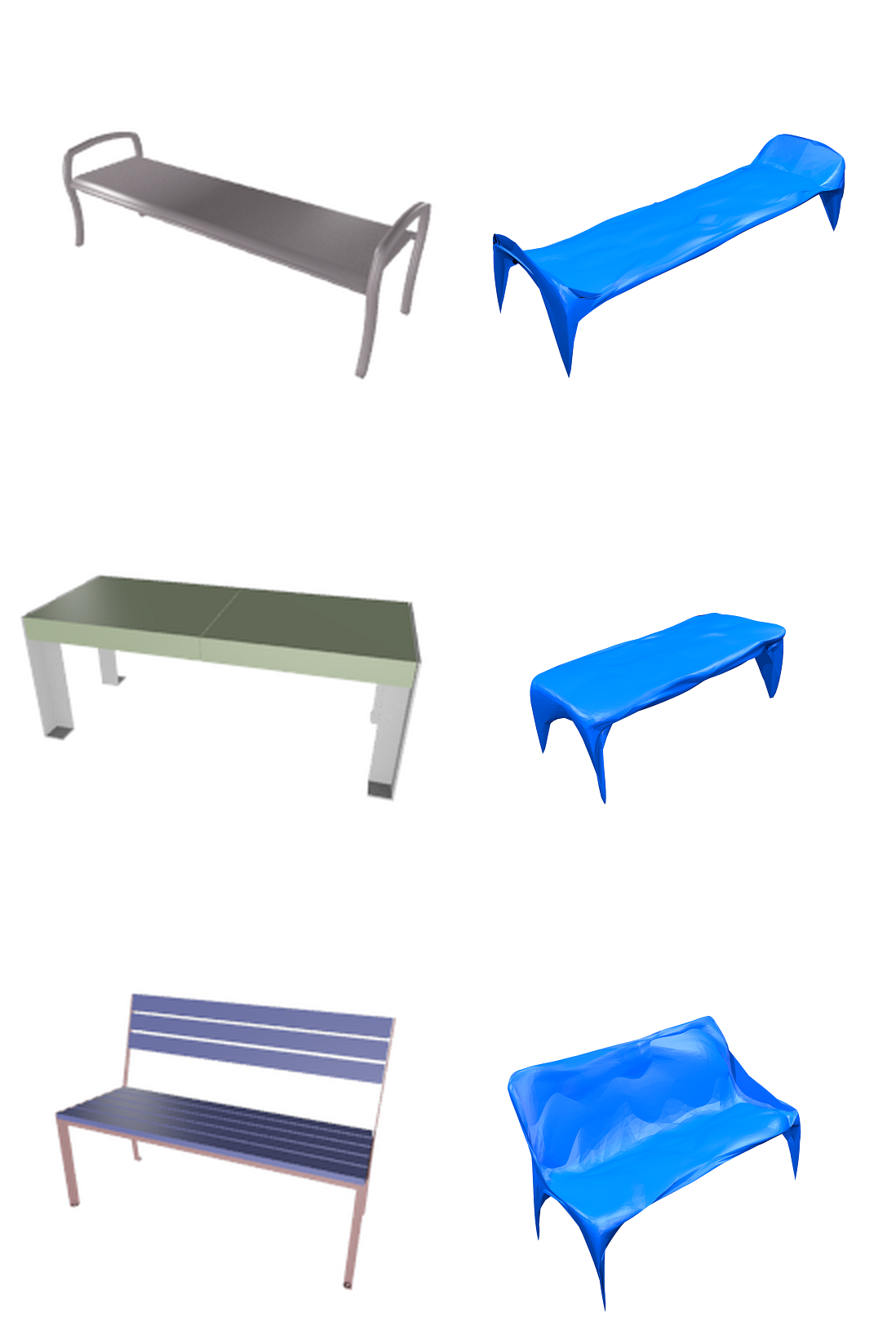}} &
\subfloat[cabinet]{\includegraphics[width = 1.5in]{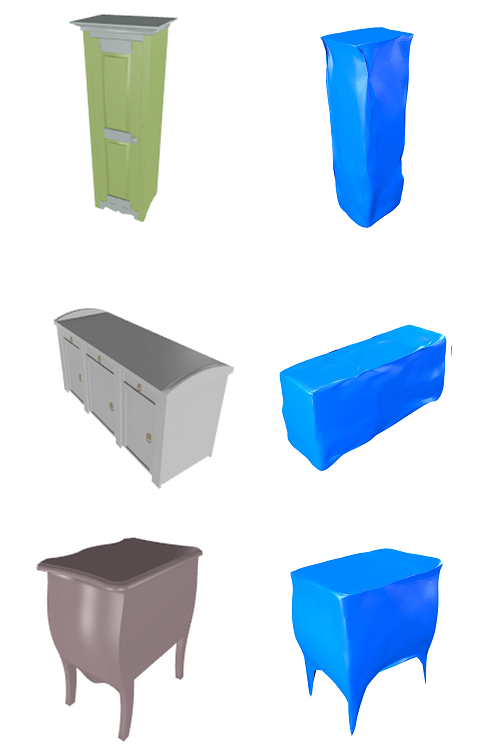}} &
\subfloat[car]{\includegraphics[width = 1.5in]{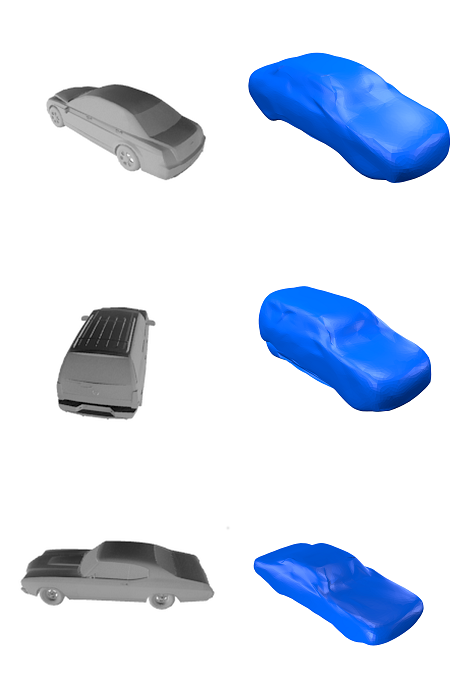}} \\
\subfloat[cellphone]{\includegraphics[width = 1.5in]{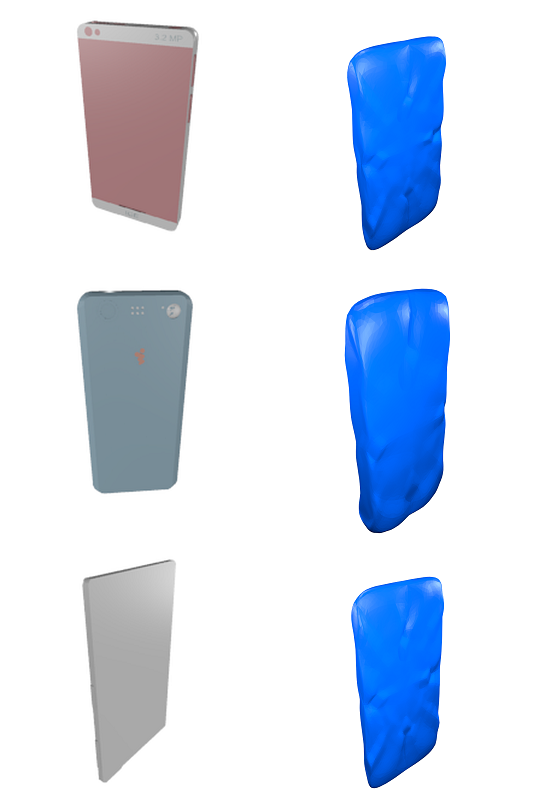}} &
\subfloat[chair]{\includegraphics[width = 1.5in]{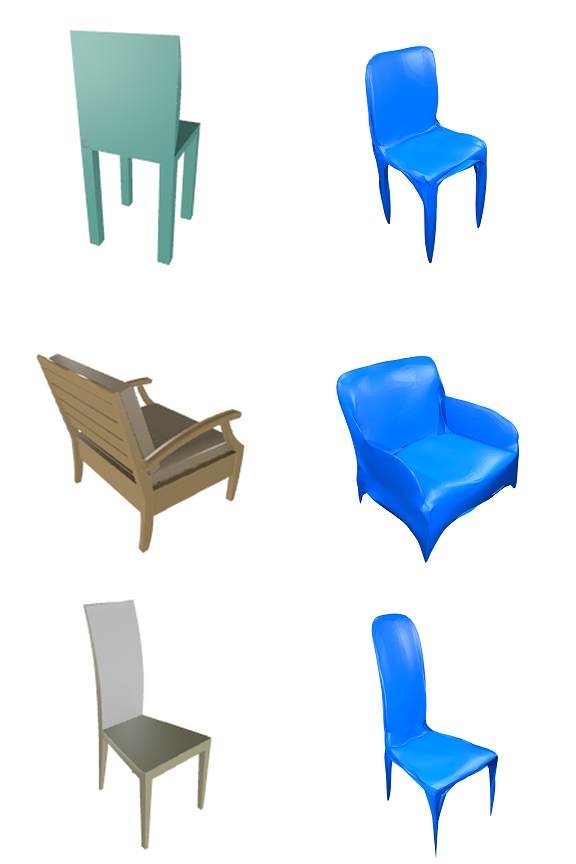}} &
\subfloat[lamp]{\includegraphics[width = 1.5in]{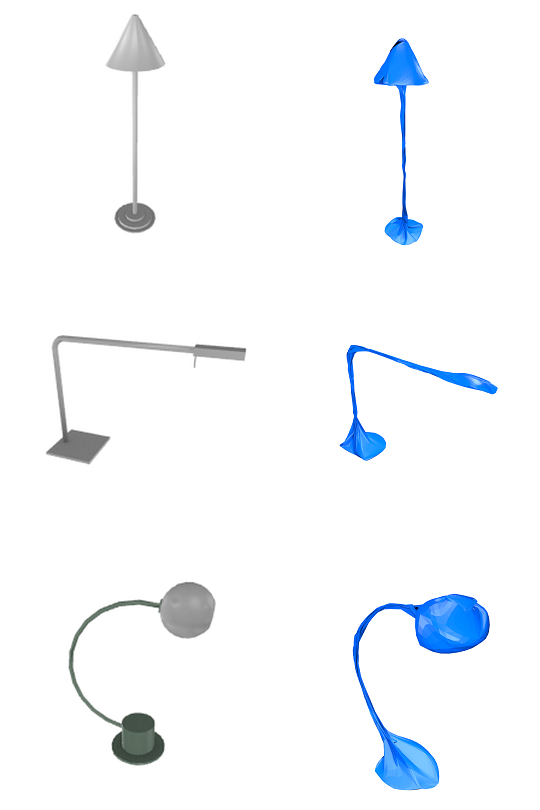}} \\
\subfloat[monitor]{\includegraphics[width = 1.5in]{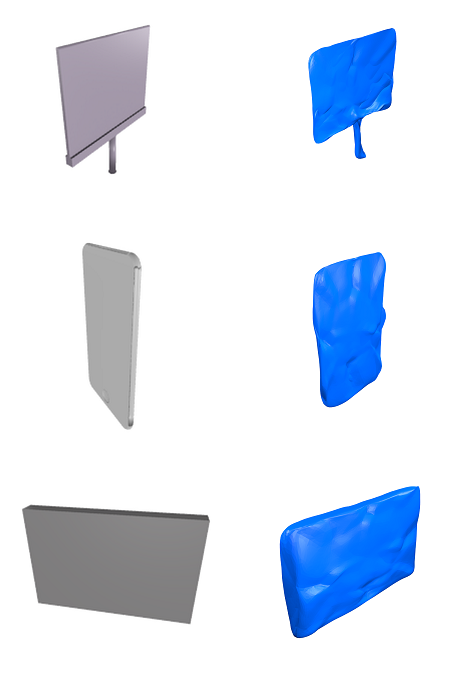}} &
\subfloat[plane]{\includegraphics[width = 1.5in]{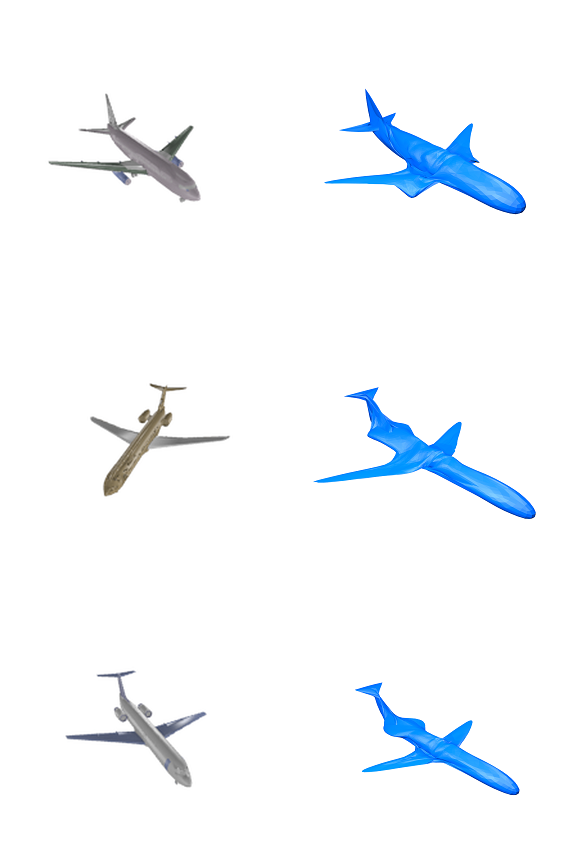}} &
\subfloat[rifle]{\includegraphics[width = 1.5in]{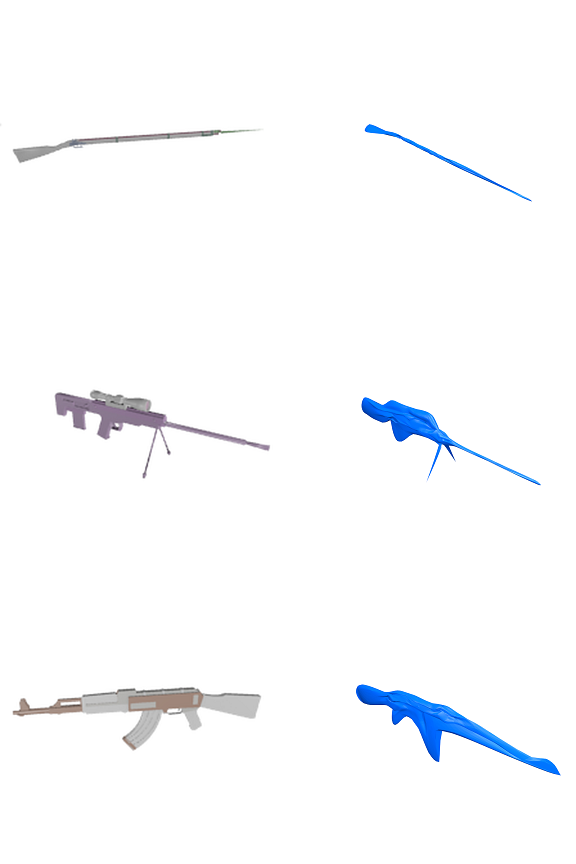}}
\end{tabular}
\caption{Single image reconstruction results on bench, cabinet, car, cellphone, chair, lamp, monitor, plane and rifle classes.}
\label{fig:recons_1}
\end{figure*}

\begin{figure*}[h!]
\centering
\begin{tabular}{cc}
\subfloat[sofa]{\includegraphics[width = 1.5in]{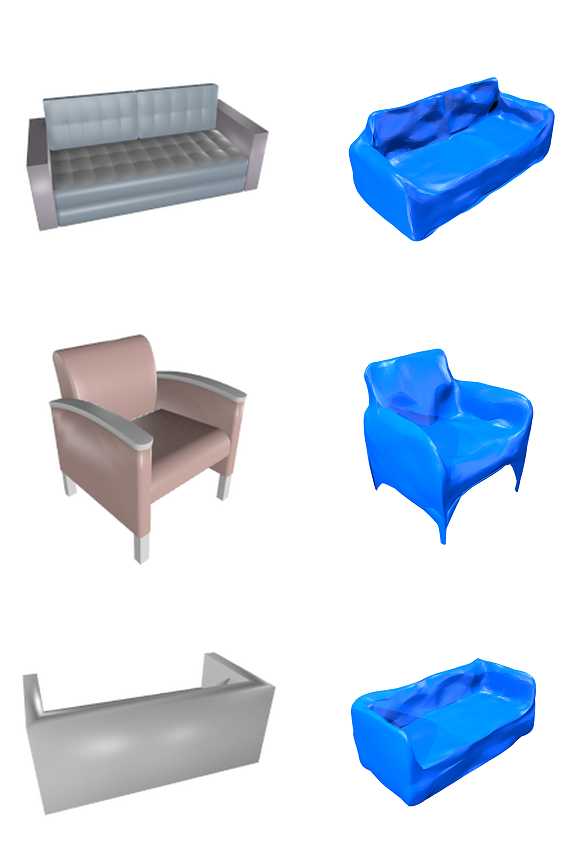}} &
\subfloat[speaker]{\includegraphics[width = 1.5in]{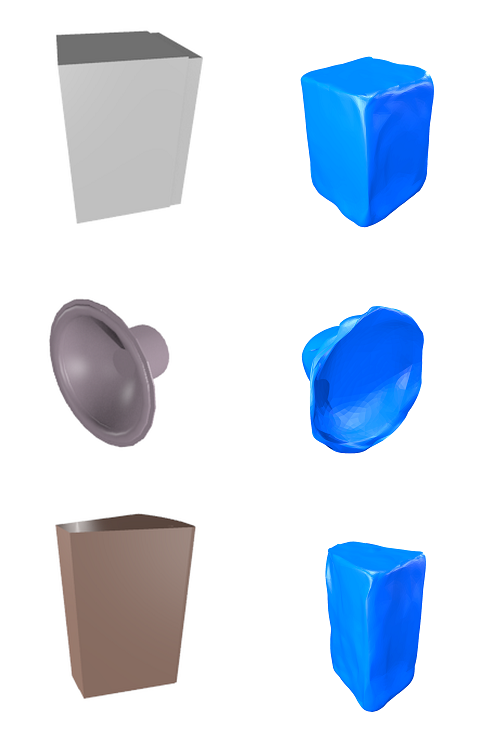}}  \\
\subfloat[table]{\includegraphics[width = 1.5in]{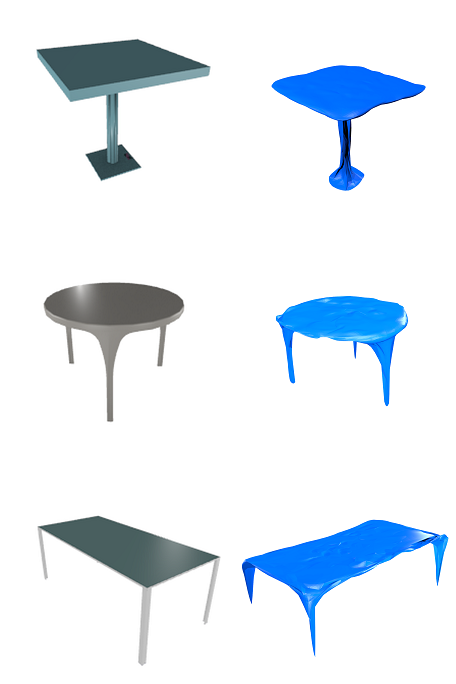}} &
\subfloat[watercraft]{\includegraphics[width = 1.5in]{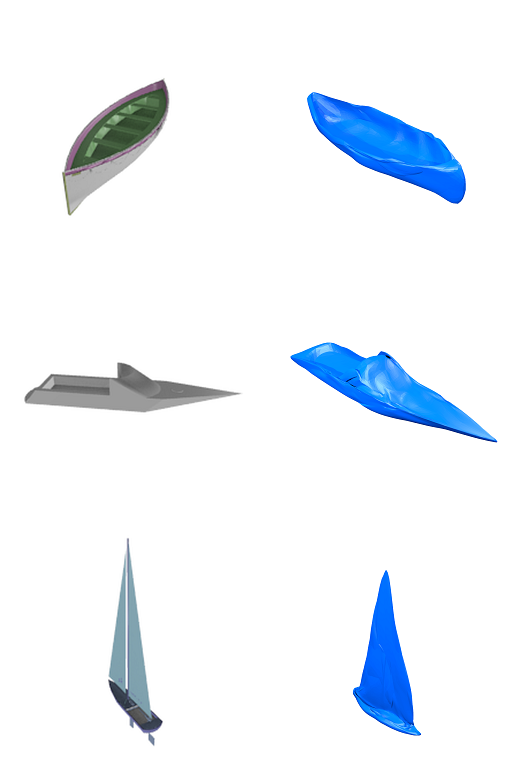}} 
\end{tabular}
\caption{Single image reconstruction results on sofa, speaker, table and watercraft classes.}
\label{fig:recons_2}
\end{figure*}

Figures \ref{fig:recons_1} and \ref{fig:recons_2} depict additional reconstruction results from each ShapeNet object class, with three objects shown per class.

\end{document}